# Leveraging AI multimodal geospatial foundation models for improved near-real-time flood mapping at a global scale

Mirela G. Tulbure[1*], Júlio Caineta[1], Mark Broich[2], Mollie D. Gaines[1], Philippe Rufin[4,3], Leon-Friedrich Thomas[5], Hamed Alemohammad[6,7], Jan Hemmerling[8], Patrick Hostert[3]

[1] North Carolina State University, Geospatial Analysis for Environmental Change Lab, Raleigh, NC, USA

[*] Corresponding author: mtulbure@ncsu.edu, orcid.org/0000-0003-1456-183X

[2] LabCorp, AI Team

[3] Humboldt-Universität zu Berlin, Earth Observation Lab, Geography Department, Germany

[4] Earth and Life Institute, UCLouvain, Place Pasteur 3, 1348 Louvain-la-Neuve, Belgium

[5] Department of Agricultural Sciences, University of Helsinki, P.O. Box 28, FI-00014 Helsinki, Finland

[6] Center for Geospatial Analytics, Clark University, Worcester, MA, USA

[7] Graduate School of Geography, Clark University, Worcester, MA, USA

[8] Thünen Institute of Farm Economics, Thünen Earth Observation, Germany

# Abstract

Floods are the most significant weather-related hazard, causing loss of life and substantial economic damage. In 2024, the warmest year on record, numerous extreme climate-driven flood events have occurred worldwide. From catastrophic flooding in Vietnam, Nigeria, and Brazil to severe storms, such as Hurricane Helene and Hurricane Milton in the southeastern USA, and flooding in Valencia, Spain, these events have had devastating impacts across five continents. Extreme rainfall has resulted in the deaths of thousands and has submerged entire towns.

Earth observation (EO) offers a valuable data source for flood mapping, thanks to its extensive coverage and frequent revisit times, with over 800 petabytes of archived data currently available. Recent advances in Geospatial Foundation Models (GFMs), such as ESA-IBM's TerraMind, enable the efficient utilization of this vast amount of data, showing promise for better generalization than traditional supervised deep learning models. However, evidence oftentimes remains inconclusive and specific to the exact downstream task and benchmark dataset used. A significant challenge for expanding the applications of supervised AI models lies in their reliance on representative labeled datasets.

Here, we fine-tune the TerraMind GFM for mapping major flood events using FloodsNet—a recent harmonized training dataset that combines spatially and temporally co-located Sentinel-1 and Sentinel-2 data for 85 global flood events—using four different backbone configurations. These configurations included two model sizes: the base (103M parameters) and the large (326M parameters) models, which differed in the number of trainable parameters, as well as two backbone configurations: frozen and unfrozen. We compare our results to the example provided by TerraMind, which fine-tunes the GFM using the Sen1Floods11 dataset. We further compared these results with those of a conventional deep learning approach, specifically the U-Net, which was trained on FloodsNet and on Sen1Floods11.

The base unfrozen configuration achieved the best balance, with strong performance across all three accuracy metrics, and with a significantly lower computational cost than the larger configuration. The best recall overall was achieved by the large unfrozen configuration. Compared to the base frozen model trained on Sen1Floods11, the model trained on FloodsNet achieved higher recall, similar overall accuracy, and lower precision. The U-Net models had better recall compared to all four GFM backbone configurations, although they yielded slightly lower overall accuracy and precision.

Integrating complementary optical-SAR data with a 10-m resolution and various AI algorithms is essential for evaluating near-real-time mapping of flood events. To our knowledge, this is among the first studies to systematically evaluate a GFM for global flood mapping across a diverse range of flood events. These findings highlight the promise of multimodal GFMs to operationalize the vast archives of EO data for climate adaptation, disaster resilience, and global risk reduction.

**Keywords:** Flood mapping; Near-real-time flood detection; SAR-optical fusion; Geospatial foundation models; Fine-tuning; Vision transformer (ViT); Self-supervised learning

# Introduction

Floods are a frequent and devastating climate-driven hazard, causing significant economic losses, displacement, and loss of life (World Meteorological Organization (WMO) 2023; Hu et al. 2018; Chen et al. 2020). Each year, up to 200 flooding events occur worldwide, and the number of flood events per year has increased over time (Guha-Sapir et al. 2016; Delforge et al. 2025). Floods have more than doubled since 2000 (Nearing et al. 2024), and the risk of flood exposure is expected to increase throughout the 21$^{st}$ century (Alfieri et al. 2017; Dottori et al. 2018) as the intensity of extreme precipitation events is projected to increase (Bender et al. 2010; Prein et al. 2017; C. Wang et al. 2022). Global economic losses since 1980 due to floods have exceeded $1 trillion (USD; Munich Reinsurance and Company 2023). While there has been a significant reduction in weather-related disaster deaths in recent decades, fatalities in low-income countries are disproportionately high (Ritchie et al. 2022; Hawker et al. 2020). Floods are increasing in frequency, amplitude, and—because of development in flood-prone areas—cost. Accurately mapping flood extent is critical for effective flood forecasting, early warning systems, and water resource management.

With its global coverage and frequent revisit times, Earth Observation (EO) data are a valuable resource for dynamic flood mapping. Most early studies of flood detection used optical, coarse-resolution imagery, such as the Moderate Resolution Imaging Spectrometer (MODIS; (Ban et al. 2017; Brakenridge and Anderson 2006; Sanyal and Lu 2004). However, the coarse spatial resolution of MODIS (250 m), and even the moderate spatial resolution of Landsat (30 m), limit the EO data's ability to detect small floods and changes in narrow water channels and along irregularly shaped shorelines (Tulbure et al. 2022; Pickens et al. 2020; Gaines et al. *under review*). While traditional methods, based on manual interpretation or rule-based algorithms, applied to EO data are helpful for local settings, they are time-consuming, often limited in accuracy, and their applicability is restricted to specific flood events in particular locations (C. Wang et al. 2022; Pastor-Escuredo et al. 2020; Levin and Phinn 2022). "Shallow" machine learning (ML) models, such as Random Forest (RF), perform well with noisy and limited training data but rely heavily on feature engineering (Breiman 2001; Maxwell et al. 2018). Such engineering requires in-depth domain knowledge and adds complexity to the design workflows, which often need to be tailored to specific regions. RFs have been used for surface water and flooding classification for Landsat (Tulbure et al. 2016), Sentinel-1 (S1; Tiwari et al. 2024), Sentinel-2 (S2; Tulbure et al. 2022; Composto et al. 2024), and PlanetScope (Hondula et al. 2021) EO data. For mapping flooding in areas with complex background signals (i.e., rice paddies or urban areas), RF classifications have yielded higher accuracies than other classification methods such as Otsu, Support Vector Machines, or even deep learning (DL) techniques (Tiwari et al. 2024; Al-Rawas et al. 2024; Flores et al. 2024). Based on supervised learning, DL does not require feature engineering; instead, it automatically extracts and refines relevant features from unstructured raw data during the training process (Munawar et al. 2022). DL models may perform better than traditional ML models (e.g., RFs) in terms of accuracy if a vast amount of high-quality training data is available (Flores et al. 2024; Wieland and Martinis 2019), which is typically non-trivial to acquire.

Most flood maps have been derived from public, optical EO data (i.e., MODIS, Landsat, and Sentinel-2; Tulbure et al. 2022; Munawar et al. 2022), but these data have several limitations, including spatial resolution, return time, and cloud obstruction (Tulbure et al. 2022; Langhorst et al. 2024). To address some of these limitations, studies have utilized Synthetic Aperture Radar (SAR) data to map floods, as it can penetrate cloud cover (Jensen et al. 2022; Feng et al. 2015; Zhao et al. 2024). SAR datasets have tradeoffs in terms of their bands (L-band is known to work better for floods; Hess et al. 1995; C. Wang et al. 2022; Zhang et al. 2016), resolution (10 m S1 vs. <1 m Capella), and slant angles. Therefore, studies are beginning to utilize multi-sensor fusion by combining SAR and optical data for flood mapping (Notti et al. 2018; Tiwari et al. 2024). Multi-sensor fusion uses data from different sensors, in our case S1 (SAR) and S2 (optical), to capture more comprehensive information for a specific area and time. The most common multi-sensor fusion technique using EO data is feature-level fusion, which can include combining spectral bands and band-derived metrics (Samadzadegan et al. 2025).

While advancements of AI techniques to derive insight from EO have enabled diverse applications (Khallaghi et al. 2023; Tuia et al. 2023; Janga et al. 2023) in domains such as agriculture (Jung et al. 2021; Rufin et al. 2024) biodiversity, conservation, climate and weather (Lv et al. 2022; Karpatne et al. 2019), and disaster response (Kuglitsch et al. 2023; Tulbure et al. 2022; Guzder-Williams and Alemohammad 2021) such as improved dynamic flood mapping (Tulbure et al. 2022; Tottrup et al. 2022; Composto et al. 2024; Tulbure et al. 2016), the biggest challenge to expanding the utility of supervised AI models is their dependency on representative and often large, labeled datasets for fine-tuning. While EO images are available globally, the heterogeneity and complexity of natural processes and human interventions worldwide necessitate regional, labeled data, which are difficult to curate.

Existing flood training datasets (i.e., WorldFloods (Mateo-Garcia et al. 2021), Sen1Floods11 (Bonafilia et al. 2020), the U.S. Geological Survey (USGS) Flood Training (Sleeter et al. 2020; Shastry et al. 2023), and the United Nations Operational Satellite Applications Programme (UNOSAT; Nemni et al. 2020)) are limited by using only one EO data modality. For example, the WorldFloods and USGS data exclusively use optical data to generate reference (e.g., water, non-water, and cloud) labels (S2 and Maxar WorldView imagery, respectively; Mateo-Garcia et al. 2021; Sleeter et al. 2020). In contrast, Sen1Floods11 and UNOSAT exclusively use SAR data (S1), and both are limited to fewer than 20 flood events worldwide (Bonafilia et al. 2020; Nemni et al. 2020). Of these four datasets, UNOSAT is the only one that distinguishes between flooded water and pre-flood water (e.g., permanent water).

The recent FloodsNet dataset (Tulbure et al. 2024a-f) harmonizes these four datasets, overcoming the sensor limitations of the individual datasets by including both optical (S2) and SAR (S1) imagery for 85 global flood events. Additionally, FloodsNet augments the reference flood data using the Global Surface Water Extent Yearly and Monthly History products (Pekel et al. 2016) to distinguish permanent, seasonal, and flood waters. Martinis et al. (2022) found that accounting for seasonal inundation was crucial in producing reliable flood maps that do not overestimate the extent of flooding.

The emergence of self-supervised learning in recent years, particularly the development of foundation models for natural image and text data, has reduced the need for large-scale labeled datasets (Bommasani et al. 2022). However, satellite imagery has unique properties compared to natural image data that need to be accounted for in the design, development, and evaluation of AI models (Rolf et al. 2024). EO data have a wide range of spatial and temporal scales, contain spectral bands beyond red, green, and blue, and the volume of the archival data is in the tens of petabytes, with an increase of tens of terabytes per day. Generation and curation of labels for these data are also expensive and, in some cases, impossible as they require being ground-referenced (Alemohammad 2021).

Self-supervised learning aims to learn lower-dimensional representations of the raw data that capture underlying patterns and causal structures, without requiring target labels during training. This learning paradigm yields a "pre-trained" model that possesses higher-level abstractions of the input data. It can thus be "fine-tuned"—typically by updating some or all of the model weights using labeled data—for specific downstream tasks, often requiring smaller labeled datasets compared to supervised learning. This approach enables the development of models for tasks and regions where collecting and curating large, labeled datasets are challenging or impossible.

Over the last couple of years, the development of geospatial foundation models (GFMs) using self-supervised learning techniques has advanced the state-of-the-art by creating task-agnostic/versatile data-driven models for geospatial applications, aiming to mitigate the issue of limited labeled data (Lu et al. 2025; Y. Wang et al. 2022). These models primarily employ Masked Autoencoder (MAE; He et al. 2021) and contrastive learning (Radford et al. 2021) as self-supervised training strategies, or adopt hybrid approaches such as Contrastive Masked Autoencoder (Huang et al. 2024; Fuller et al. 2023), focusing on various modalities of satellite imagery as input, including optical and SAR data. Self-supervised learning techniques can be applied to vast amounts of unlabeled satellite imagery to develop GFM that can be fine-tuned with limited labeled data for tackling a variety of tasks in diverse geographic contexts (Mo et al. 2023; Lacoste et al. 2021).

Since the release of the Vision Transformer (ViT) architecture, over 60 GFMs have been developed for satellite imagery (Lu et al. 2024; Ali Braham et al. 2025; Stewart et al. 2023; Wang et al. 2023), spanning a mix of convolutional and transformer-based architectures. However, the rise in the development of GFMs has not advanced the science of task-specific applications (Ghamisi, Yu, Zhang, et al. 2025; Xie et al. 2024), such as models for dynamic flood mapping. To address this shortcoming, this research aims to tackle the challenges of dynamic flood mapping by leveraging the power of pre-trained multi-sensor GFMs, specifically by fine-tuning a newly released multi-sensor/multimodal foundation model, TerraMind (TM; Jakubik et al. 2025), to the task of dynamic flood mapping using the FloodsNet dataset (Tulbure, et al. 2024a-f), a recently available harmonized dataset of S1 and S2 data. Similar to the original TM paper, we benchmark our fine-tuned models against a conventional supervised learning strategy, a U-Net baseline model. This not only ensures that the GFM models are benchmarked against a widely used architecture but also enables further investigation of how GFMs compare against supervised models in specific downstream tasks.

**Methods**

FloodsNet. We trained and evaluated all models using FloodsNet (Tulbure, et al. 2024a–f), a harmonized dataset of 85 global flood events designed for multi-sensor training of flood extent models. FloodsNet integrates and co-registers S1 SAR and S2 MSI data, combining existing open datasets—WorldFloods (Mateo-Garcia et al. 2021), Sen1Floods11 (Bonafilia et al. 2020), USGS Flood Events (Sleeter et al. 2020), and UNOSAT Flood Maps (Nemni et al. 2020)—and augments them with corresponding missing sensor data and seasonal and permanent water occurrence masks from the JRC Global Surface Water dataset (Pekel et al. 2016). The dataset was used to train and validate all U-Net models and to fine-tune the GFM models in this study. For this research, we used FloodsNet without UNOSAT because its significantly larger raster sizes would lead to an uneven distribution of data sources across the data splits, potentially introducing bias. Furthermore, the absence of a cloud mask in UNOSAT, for the corresponding S2 data, would lead to the models receiving false information. Still, we will refer to FloodsNet throughout the remainder of this paper, where we used WorldFloods, Sen1Floods11, and USGS.

GFMs. We used TM, a newly released GFM developed by IBM and ESA-Φ lab (Jakubik et al. 2025), because it is a multimodal model trained on multi-sensor data from S1 and S2. TM was shown to generally perform better than other GFM and U-Net models when using PANGEA, a community-standard benchmark for EO data (Jakubik et al. 2025). We used the TerraTorch (Kienzler et al. 2025) framework to fine-tune the TM foundation model. TerraTorch simplifies fine-tuning of GFMs by integrating and standardizing reusable backbones with state-of-the-art decoders and heads. The library provides easy access to open-source pre-trained GFMs backbones, including TM.

We ran the TM GFM using both the base (terramind_v1_base) and large (terramind_v1_large) backbones, with frozen and unfrozen backbones, resulting in a total of four configurations. All models were trained on FloodsNet using both S1 (S1GRD, using the VV and VH polarizations) and S2 (S2L1C, using all 13 bands) data as inputs. We employed a U-Net-style decoder for binary water segmentation. We optimized the models using the AdamW optimizer with a learning rate of 2e-5 and a Dice loss function. Training was conducted over 20 epochs with a batch size of 8, using per-modality normalization statistics computed from the dataset (as provided in the TM example). The best model was automatically selected based on the lowest validation loss during training.

All training hyperparameters (optimizer, learning rate, batch size, loss function) were adopted directly from the TM example, except for the number of epochs, which we increased to 20. Aside from substituting the training dataset—using FloodsNet in place of Sen1Floods11—no further hyperparameter tuning or search was conducted. This was a deliberate choice, as the focus of our study was to isolate and evaluate the impact of different backbone configurations (base vs. large, frozen vs. unfrozen) and training datasets, rather than to optimize model performance.

To train the GFMs, we used Google Colab Pro with a T4 GPU, except for the large unfrozen configuration, for which we utilized an A100 GPU. At the time of analysis (May–July 2025), the usage costs per GPU type in Google Colab were 6.86 computing units per hour for the A100, 1.64 units per hour for the T4, and 2.09 units per hour for the L4.

We evaluated TM by comparing its performance on FloodsNet vs. Sen1Floods11—the example dataset used in TM for flood mapping—using four different backbone configurations. These configurations included two model sizes: the base (103M parameters) and the large (326M parameters), which differed in the number of trainable parameters, as well as two backbone configurations: frozen and unfrozen. In the frozen configuration, the backbone's parameters are kept fixed during training, and only the additional layers on top (e.g., task-specific heads) are trained. In contrast, in the unfrozen configuration, all the weights of the backbone (pretrained model) are updated during training. We used the same hyperparameters (as described above) across the four configurations.

The evaluation was done with identical test sets. In Table 1, four models (representing the four different backbone configurations) were trained on FloodsNet. Then, each of the four models was tested on two test sets: the test split from FloodsNet and the test split from Sen1Floods11 (three examples shown in Figure 1). In fact, the test split from Sen1Floods11 is a subset of the test set from FloodsNet (i.e., FloodsNet also includes Sen1Floods11). This was done to allow the comparison with the default example from TM, which only used the Sen1Floods11 data.

**U-Net**. We also assessed the performance of two U-Net neural network (Ronneberger et al. 2015) models—one trained on FloodsNet, and the other trained on Sen1Floods11—using the same training split as in the TM models (FloodsNet and Sen1Floods11), and with the same 15 bands as inputs (2 from S1 and 13 from S2) and 100M learnable parameters. We implemented the U-Net models in PyTorch and trained them on Google Colab Pro using the L4 GPU. We conducted the training for 50 epochs with a batch size of 32, using the AdamW optimizer (learning rate 1e-4). The loss function was focal loss, with class-balancing weights (alpha) set as the inverse of class frequencies and gamma fixed at 2. All inputs were clipped to sensor-specific ranges (SAR: [−50, 25] dB; optical reflectance: [0, 1]). The best-performing model was selected based on the lowest validation loss during training. The two U-Net models were then tested on the same test splits as in the TM models (FloodsNet and Sen1Floods11).

**Results:**

Comparison of TM across the four configurations. The base unfrozen configuration trained on FloodsNet achieved the best balance, with strong performance across all three metrics (Table 1, rows 3 and 4; output examples shown in Figure 1), and without incurring a significantly higher computational cost (+44 min on T4 GPU vs. the base frozen configuration). The best recall overall was achieved by the large unfrozen configuration (91.37% and 91.39%, tested on FloodsNet and Sen1Floods11, respectively) trained on FloodsNet. This was the only model trained on an A100 GPU (for 157 minutes; 4.86 times the cost of training the large unfrozen backbone), whereas all other configurations ran on a T4 GPU.

**Table 1.** Performance metrics per TerraMind backbone configuration. The four models were trained on FloodsNet and tested on FloodsNet vs. Sen1Floods11. The model in the TerraMind Example (last two rows) was trained on Sen1Floods11 and tested on FloodsNet vs. Sen1Floods11. Values in bold are the best accuracy metrics achieved. Precision and recall refer to the class of interest ("water").

| Backbone | Test | Accuracy | Precision | Recall |
|---|---|---|---|---|
| Base frozen | FloodsNet | 96.53 | 77.67 | 87.74 |
|  | Sen1Floods11 | 97.12 | 88.57 | 88.07 |
| Base unfrozen | FloodsNet | 96.95 | 80.13 | 89.06 |
|  | Sen1Floods11 | **97.51** | 90.58 | 89.19 |
| Large frozen | FloodsNet | 96.33 | 75.61 | 89.12 |
|  | Sen1Floods11 | 97.12 | 87.79 | 89.11 |
| Large unfrozen | FloodsNet | 96.67 | 76.96 | 91.37 |
|  | Sen1Floods11 | 97.35 | 87.70 | **91.39** |
| TerraMind Example (base frozen) | FloodsNet | 96.79 | 81.21 | 84.96 |
|  | Sen1Floods11 | 97.10 | **90.98** | 85.04 |

Unfreezing the backbone improved performance, although not significantly. This improvement was observed for both the base and the large backbones, and regardless of test split. The largest improvement was in precision in the base model (+2.46%pt increase in base unfrozen compared to base frozen, when tested on FloodsNet; Table 1, rows 3 and 1), which also increased training time from 113 to 157 min (+44 min, +39%, base backbone). The second largest improvement was in recall, when comparing the large frozen configuration to the large unfrozen configuration, on both test sets (Table 1; +2.28% on Sen1Floods11, rows 8 and 6; and +2.25%pt in FloodsNet, rows 7 and 5).

The larger backbone did not consistently outperform the base backbone, with the large un/frozen model vs. the base un/frozen model showing an improvement in recall; however, we noted a drop in overall accuracy and precision, on both test splits (Table 1).

Compared to the base frozen model trained on Sen1Floods11 (Table 1, last two rows), the model trained on FloodsNet (Table 1, first two rows) achieved higher recall (+2.78%pt and +3.03%pt), similar overall accuracy (-0.26%pt and +0.02%pt), and lower precision (-3.54%pt and -2.41%pt). Percent point differences in parentheses refer to testing in FloodsNet and Sen1Floods11, respectively.

We split our FloodsNet into training, validation, and test sets, and trained the models on the training subset. To avoid data leakage, the dataset was organized by flood event before splitting, such that no image from the same flood event is present in more than one data split. We tested and compared the four configurations on the test holdout from FloodsNet, as well as the test holdout from Sen1Floods11 used in TM, to make the testing part as similar as possible. We found that when testing on the Sen1Floods11 (default), the results were consistently better than when testing on the FloodsNet holdout. The FloodsNet test holdout used images from 15 flood events from around the world. The test holdout included the Sen1Floods11 test data used in TM (10 flood events) and data from USGS (2 flood events) and WorldFloods (3 flood events). The Sen1Floods11 test split had lower average (12.1%) and median (0.1%) no-data/cloud-per-image ratios than the FloodsNet test split—which includes Sen1Floods11—(average 12.7%, median 0.8% for no-data/cloud). The Sen1Floods11 test split had higher average (10.6%) and median (1.7%) water-per-image ratios than the remaining FloodsNet test split (2.8% and 1.9% average and median, respectively). These differences suggest that the Sen1Floods11 test data alone consisted of images that were easier for the model to predict than the full FloodsNet test data.

**Table 2.** Performance metrics for the U-Net. The two models were trained on FloodsNet and on Sen1Floods11. Then, each model was tested also on FloodsNet and Sen1Floods11. Both the training and test splits are the same as in the TM models. Values in bold are the best accuracy metrics achieved. Precision and recall refer to the class of interest ("water").

| **Models** | **Test** | **Accuracy** | **Precision** | **Recall** |
|---|---|---|---|---|
| Sen1Floods11 | FloodsNet | 96.16 | 72.51 | **94.08** |
|  | Sen1Floods11 | **97.62** | 87.91 | 93.65 |
| FloodsNet | FloodsNet | 96.71 | 77.31 | 91.10 |
|  | Sen1Floods11 | 97.55 | **89.42** | 91.01 |

**Comparison with the U-Net**. The U-Net achieved similar overall accuracy to the four TM backbone configurations (Table 1), with slightly lower precision (Table 2). However, it produced the highest recall across all models (94.04; Table 2, row 1), with an increase of +2.69%pt compared to the best-performing GFM configuration (91.39; Table 1, row 8). When both training and testing were performed on FloodsNet, the U-Net achieved an overall accuracy of 96.71%, precision of 77.31%, and recall of 91.10% (Table 2).

When trained on FloodsNet and tested on the Sen1Floods11 dataset, the U-Net achieved an accuracy of 97.62%, precision of 87.91%, and recall of 93.65%, outperforming all GFM configurations in recall (Table 1). In comparison, the highest recall among GFMs was 91.39%, achieved by the large unfrozen backbone (Table 1).

The highest precision across all models was achieved by the base unfrozen GFM, with a value of 90.98% when trained and tested on Sen1Floods11. The U-Net's highest precision, 89.42%, was slightly lower. Across datasets, GFMs showed higher precision and overall accuracy, whereas the U-Net consistently produced higher recall.

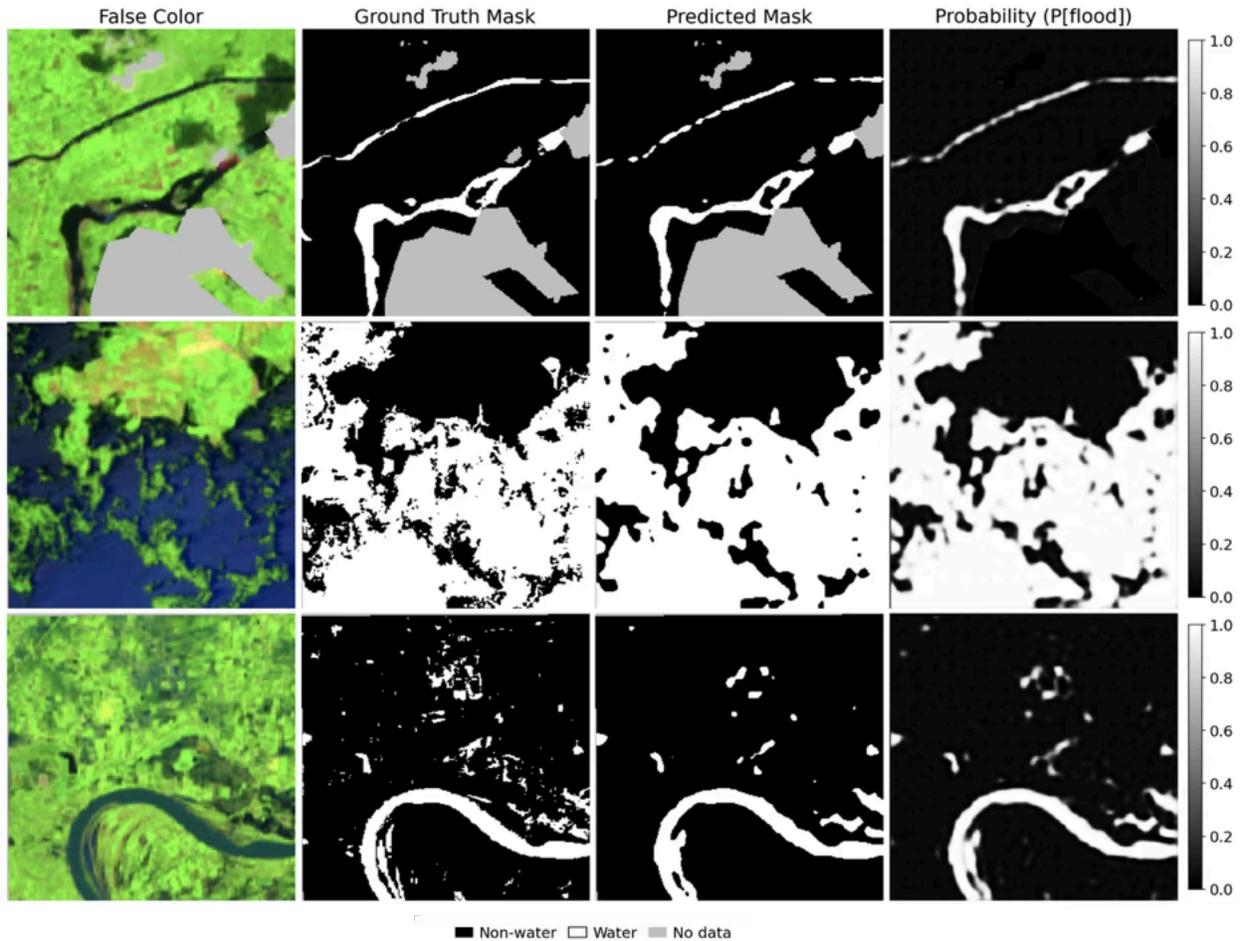

*Figure 1*. *Predicted flood maps using the TerraMind base backbone unfrozen configuration and fine-tuned on FloodsNet. The floods are from the Sen1Floods11 test subset: (top) India, 2016-06-29; (middle) Cambodia, 2018-07-17; and (bottom) USA, 2019-05-20. The first column shows a false color image from Sentinel-2 data (R: band 11, SWIR1; G: band 8A, Narrow NIR; B: band 4, Red), the second column consists of ground truth data with water shown in white and non-water in black, the third column shows the predicted flood maps (no data shown in grey), and the last column shows the estimates of the model confidence (softmax probabilities). The*

*probability values are scaled down using a fixed temperature parameter of T=2. This parameter only adjusts the visualization of the probabilities, without changing the predicted class. The model's raw probabilities (T=1) are sharply peaked, showing that the model is over-confident in classifying either class (water/non-water).*

**Flood maps.** Visually, the predicted flood maps generated using the TM fine-tuned on FloodsNet (Figure 1, third column, "Predicted Mask") closely matched the original images (Figure 1, first column, "False color") and the provided ground-truthed flood event labels (Figure 1, second column, "Ground Truth Mask"). However, the GFM fine-tuned on FloodsNet produced smooth, contiguous flood patches ('blobby' patterns), which differ from the more fragmented appearance typical of pixel-wise models (Zhou et al. 2020). In general, the TM models correctly classified the main flooded area in the input dataset, including rivers (example in Figure 1, top and bottom rows) and lakes (Figure 1, bottom row). The models were also capable of avoiding the misclassification of cloud shadow in the absence of corresponding labeled data in the input (example in Figure 1, top row). The visualization of the softmax probabilities (Figure 1, last column) shows that the models were overconfident in either the true and false positives predictions. This behavior is possibly enforced by the nature of a binary prediction task. However, it is still possible to see that some false negatives (missed predictions) show up with a probability just under 0.5 (for example, in the top corner of the bottom row in Figure 1).

**Discussion:** GFMs are trained on large-scale unlabeled data; however, their applicability to specific downstream tasks depends on the model's ability to generalize beyond the original distribution of the training data. The promise of GFMs is that they generalize better than supervised DL models (Khallaghi et al. 2024), although the evidence remains inconclusive for some downstream tasks (Xie et al. 2024; Ghamisi, Yu, Marinoni, et al. 2025). This is because while many GFMs do well on benchmark datasets, few are deployed in real-world systems, where data heterogeneity is higher. This is particularly the case with large language models (LLMs), where it has been found that while certain LLMs can perform better on benchmark datasets, with more training, they perform poorer on downstream tasks (Nagarajan et al. 2025). We hypothesized that, given the limited available labeled data for dynamic flood mapping, GFMs will outperform supervised models (i.e., U-Net) for this task. To test this hypothesis, we used training and validation data from the FloodsNet dataset (Tulbure, et al. 2024a–f). We evaluated the performance of the TM GFM fine-tuned with FloodsNet and tested its performance both on the FloodsNet and Sen1Floods11 (as in the TM example) test holdouts for a fair comparison. We further compared it against the performance of U-Net models trained on FloodsNet and on Sen1Floods11.

The performance differences among the four TM configurations largely align with trends reported in other GFM studies. In our case, unfreezing the backbone yielded modest gains, particularly in precision for the base model (+2.46%pt; Table 1, rows 3 and 1), which is consistent with expectations, as unfreezing enables the model to more fully adapt to the task-specific flood data during fine-tuning. Interestingly, while the large unfrozen configuration achieved the highest recall (91.37% on the FloodsNet test split, and 91.39% on the Sen1Floods11 test split; Table 1, rows 7 and 8), it did not deliver consistent improvements in overall accuracy or precision compared to the base unfrozen model (Table 1, rows 3 and 4). This finding aligns with observations in the Prithvi-EO-2.0, where the 600M-parameter version showed a higher recall than the 300M version (and earlier Prithvi-1.0), in detriment of a lower precision (Szwarcman et al. 2025). Similarly, fine-tuning the TM (frozen base backbone) in the larger dataset—i.e., FloodsNet vs Sen1Floods11—also resulted in higher recall at the cost of a lower precision (Table 1; first two and last two rows).

The lack of a clear performance advantage for the large backbone in our study may also be attributed to task complexity and dataset characteristics. Flood mapping is a challenging problem with imbalanced classes (flood pixels being much rarer than non-flood pixels), and larger models are more sensitive to these imbalances. Our results suggest that for operational flood mapping, the base unfrozen configuration offers the best trade-off, achieving strong accuracy and precision without substantially increasing computational costs. In contrast, the large unfrozen configuration may be advantageous only when maximizing recall is the priority, such as for rapid disaster response.

The fact that the U-Net performed slightly worse (Table 2) than TM in terms of overall accuracy and precision is consistent with the expected advantages of GFMs, which benefit from pretraining on massive amounts of data and therefore exhibit strong generalizability (Li et al. 2023). However, the U-Net achieved higher recall compared to all four TM configurations, a result that has also been reported in other studies, and a key advantage in flood mapping, where undetected inundation can compromise quick response efforts. For instance, a recent evaluation of burn intensity mapping found that a U-Net performed similarly to a GFM (Prithvi) in terms of mean IoU and mean F1 score (Szwarcman et al. 2025). Likewise, Li et al. (2023) observed that U-Net achieved higher recall on in-domain test data for flood mapping, whereas the GFM generalized better to unseen regions, highlighting the complementary strengths of the two approaches: U-Net's ability to capture fine spatial details such as narrow river channels and urban flooding, and GFM's ability to transfer knowledge across diverse geographies and conditions (Kostejn et al. 2025). Furthermore, in event-specific applications, U-Net has achieved recalls exceeding 90% when mapping individual flood events using S1 data (Fakhri and Gkanatsios 2025).

While the fine-tuned GFM flood predictions closely matched the reference imagery and ground truth, the maps exhibited a smoother, 'blobby' texture (Figure 1). Such patterns are common in context-aware DL approaches, which emphasize spatial coherence and region-level consistency over pixel-level sharpness (Zhou et al. 2020). Architecturally, Vision Transformer-based GFMs tend to favor global context and generalization across diverse geographies, but they can miss fine-grained, small-scale features (Kostejn et al. 2025). In our case, this effect may also be partly influenced by the fine-tuning configuration (e.g., number of epochs, learning rate) and reflects the inherent precision–recall trade-offs of flood mapping. Although these smoother outputs reduce boundary sharpness, they also suppress the salt-and-pepper noise typical of purely pixel-wise models, leading to more coherent flood extents and often higher recall—a key priority when mapping rare, high-impact events such as floods (Bentivoglio et al. 2022). This balance between detail and coherence is crucial for operational flood mapping, where the cost of missing flooded pixels is typically higher than that of overprediction.

GFMs that integrate multimodal data from complementary sensors, such as S1 (SAR) and S2 (optical), provide a comprehensive basis for flood mapping. For instance, SSL4EO-S12 (Wang et al. 2023) demonstrated the benefits of combining SAR and optical data streams for a range of EO tasks. Multi-sensor integration is particularly important for ephemeral flood detection, where relying on a single data source often misses short-lived or cloud-obscured events (Tulbure et al. 2022). Previous studies have also shown that combining these complementary data streams improves flood segmentation accuracy by leveraging complementary spectral information (Shastry et al. 2023; Tamura-Wicks et al. 2025).

While other GFMs are capable of multimodal integration (e.g., Clay, https://madewithclay.org/), the TM model used here is well-documented and straightforward to implement. Other GFMs have been fine-tuned specifically for flood segmentation but rely solely on optical data, which can be problematic during flood events due to cloud cover. For example, Prithvi-EO-2.0 achieved strong performance, with an F-score of 97% and an IoU of 83% (Szwarcman et al. 2025). However, models like TM, which natively integrate SAR and optical data, are better suited for operational flood mapping, especially in challenging conditions where optical imagery alone is insufficient.

By developing more accurate AI models using limited labeled data for mapping flooding dynamics, we aim to enhance resilience to extreme weather events. Accurate dynamic flood mapping provides critical real-time flood maps that help identify and enable timely flood warnings, allowing government agencies and citizens to take necessary precautionary measures. AI models, particularly self-supervised GFMs, can enhance the application of satellite imagery for flood mapping. However, given the prevalence of GFMs in recent years, it is clear that choosing a foundation model and fine-tuning it for image classification or semantic segmentation tasks is paramount.

Our research also enhances water resource management, as precise water extent data can enable informed decision-making for water allocation strategies and drought mitigation efforts. Last, our study contributes to the broader field of developing and utilizing GFMs for geospatial tasks, a priority for federal agencies (e.g., the NASA Science Mission Directorate). By producing more accurate and timely flood maps, our models support rapid identification of affected areas and enable faster, more targeted disaster response. This capability allows emergency managers to prioritize resources, coordinate evacuation efforts, and mitigate the human and economic impacts of extreme events. Beyond the immediate response phase, these models also play a key role in post-disaster assessment, including estimating damage and tracking the progress of recovery over time. By integrating data from multiple sources, they provide a comprehensive view of flooding dynamics, improving decision-making for both short-term actions and long-term mitigation and adaptation planning.

GFMs have the potential to reduce the computational and data requirements for producing accurate flood maps, accelerating the transition from research to operational use. Similar to the early DL trajectory around 2016, we are currently seeing incremental improvements in accuracy with GFMs, but limited attention to their societal, environmental, and economic implications (Ghamisi, Yu, Zhang, et al. 2025). As others have argued (Ghamisi, Yu, Zhang, et al. 2025), this calls for a paradigm shift from a technology-centric focus on model development to an impact-driven framework that prioritizes downstream deployment and decision-making—widely regarded as the ultimate goal of Responsible AI in EO (Ghamisi, Yu, Marinoni, et al. 2025). However, while broad frameworks are valuable, there is also a critical need for task-specific evaluations that integrate expert knowledge and domain expertise. For example, for flood mapping, the priority should not be developing a model that also performs reasonably well on unrelated tasks such as field boundary detection. Instead, evaluation should go deeper into flood-specific performance and uncertainty, ensuring the outputs are reliable and actionable for disaster response and risk management. This targeted approach complements broader GFM benchmarking by aligning evaluation with the operational realities of high-stakes applications, such as flood mapping. Ultimately, advancing Responsible AI in EO will require linking these two levels of evaluation, ensuring that models are both broadly robust and validated for the specific societal challenges they are deployed to address.

**Acknowledgements**. This work was supported by a NASA Terrestrial Hydrology Project (Grant Number 80NSSC21K0980). MGT was partly supported by an Alexander von Humboldt Fellowship for experienced researchers.

**Data and code availability**. The FloodsNet dataset (Tulbure, et al. 2024a–f) used for model training and validation was developed by harmonizing four open-access datasets (WorldFloods, Sen1Floods11, USGS Flood Events, and UNOSAT Flood Maps). Upon publication, FloodsNet will be released under an open license, and its DOI will be provided. The derived model predictions and code for fine-tuning the TM geospatial foundation model are available from the corresponding author upon reasonable request.

**Author contribution**: MGT designed the study, implemented analyses, wrote the first draft, acquired the funding, and supervised the project. JC implemented most analyses and provided edits to the draft. All other co-authors read and edited the draft.